\documentclass[runningheads]{llncs}

\usepackage[review,year=2024,ID=*****]{eccv}
\usepackage{eccvabbrv}

\usepackage{graphicx}
\usepackage{booktabs}

\usepackage[accsupp]{axessibility}  % Improves PDF readability for those with disabilities.

\usepackage[pagebackref,breaklinks,colorlinks,citecolor=eccvblue]{hyperref}
\usepackage{orcidlink}
\usepackage[misc]{ifsym}

\begin{document}
\nolinenumbers
\title{Facial Affect Recognition based on Multi Architecture Encoder and Feature Fusion for the ABAW7 Challenge} 

\titlerunning{Abbreviated paper title}

% TODO FINAL: Replace with your author list. +
\author{Kang Shen\inst{1} \and Xuxiong Liu\inst{1} \and
Jun Yao \and Boyan Wang  \and Yu Wang \and Yujie Guan \and Xin Liu \and Gengchen Li \and
Xiao Sun\Letter}


\institute{Hefei University of Technology\\
Hefei, China\\
\email{\{shenkang,liuxuxiong,  yaojun, wangboyan,2023170653\}@mail.hfut.edu.cn}\\
\email{\{320522038,2312789664,\}@mail.hfut.edu.cn,ligengchen6599@163.com} \\
\email{sunx@hfut.edu.cn}}

\maketitle

\begin{abstract}
  In this paper, we present our approach to addressing the challenges of the 7th ABAW competition. The competition comprises three sub-challenges: Valence Arousal (VA) estimation, Expression (Expr) classification, and Action Unit (AU) detection. To tackle these challenges, we employ state-of-the-art models to extract powerful visual features. Subsequently, a Transformer Encoder is utilized to integrate these features for the VA, Expr, and AU sub-challenges. To mitigate the impact of varying feature dimensions, we introduce an affine module to align the features to a common dimension. Overall, our results significantly outperform the baselines.

\end{abstract}

\section{Introduction}
\label{sec:intro}
Sentiment analysis, as a key area in pattern recognition, is driving human-computer interaction to a deeper emotional dimension. Sentiment behavior analysis, through in-depth study of non-verbal signals such as human facial expressions, speech and body language, aims to improve the emotion understanding ability of AI systems and achieve more natural and empathetic human-computer communication. With the deepening research on sentiment behavior analysis in natural scenarios, we are gradually developing more advanced AI technologies, and these play a key role in a variety of fields such as technical human-computer interaction, healthcare, driving safety, advertising optimization, education and learning, and virtual reality.

Facial Expression Recognition (FER) shows great potential in the aforementioned areas, and despite significant advances in facial recognition technology, the fine-grained nuances of emotion understanding are still an unsolved challenge. Facial information is the most intuitive and real element of emotional expression, therefore, we focus on facial emotion recognition and comprehensively present a solution aimed at addressing three major challenges of affective behavior analysis in a wild environment (ABAW): first, valence-arousal (VA) estimation, i.e., accurately determining positive and negative emotions and their activation levels; and second, expression (Expr) recognition, which is aimed at identifying, e.g., six basic (Anger, Disgust, Fear, Happiness, Sadness, Surprise) and Neutrality; and third, Action Unit (AU) Detection, which analyzes facial muscle movements to capture subtle facial gestures and then decode complex emotional expressions. This comprehensive solution is dedicated to improving the accuracy and utility of sentiment analysis.

Based on the rich multimodal nature of the Aff-Wild 2 dataset, we are committed to mining the sentiment information embedded in it and enhancing the real-world utility of the analysis method. To achieve this goal, we adopt three key strategies. First, a self-supervised Masked Auto Encoder (MAE) model is utilized to learn high-quality sentiment feature representations from large facial image datasets to optimize subsequent task performance. Second, through the Transformer-based model architecture, we effectively fuse information from multimodal data to facilitate the interaction of different modalities such as audio, video, and text. Finally, using an integrated learning strategy, we subdivide the dataset into multiple sub-datasets, assign them to different classifiers, and integrate the outputs of these classifiers in order to obtain more comprehensive and accurate sentiment analysis results.

Our proposed method is innovative and breakthrough in three aspects:

(1) We integrate and optimize a large-scale facial expression dataset, and through fine-tuned processing, we successfully construct an efficient facial expression feature extractor, which significantly improves the performance of our model on specific facial expression recognition tasks.

(2) We introduced a Transformer-based multimodal integration model to address the three sub-challenges of the ABAW competition. This model effectively facilitates complementarity and fusion between different modal data, thus enhancing the extraction and analysis of expression features.

(3) We adopted an integrated learning strategy to address the need for sentiment analysis in different scenarios. This approach trains classifiers independently on multiple sub-datasets with unique contexts, and improves the overall accuracy and generalization ability by integrating the prediction results of these classifiers, which ensures the stable operation of our model in diverse application scenarios.

\section{Related Work}
\label{sec:relat}

\subsection{Multimodal Features}

In prior editions of the  ABAW\cite{zafeiriou2017aff,kollias2019face,kollias2019expression,kollias2020analysing,kollias2021affect,kollias2021analysing,kollias2021distribution,kollias2022abaw,kollias2023abaw,kollias2023abaw2,kollias20246th,kollias2023multi,kollias2019deep,kollias20247th}, a plethora of multimodal features, encompassing visual, auditory, and textual characteristics, have been extensively employed. By extracting and analyzing these multifaceted features, the performance of affective behavior analysis tasks can be significantly enhanced.

In the visual modality, facial expressions constitute a crucial aspect for understanding and analyzing emotions. The human face is represented by a specific set of facial muscle movements, known as Action Units (AUs)\cite{martinez2017automatic}, which have been widely adopted in the study of facial expressions.

Within the context of affective computing, auditory features—typically encompassing energy characteristics, temporal domain features, frequency domain features, psychoacoustic features, and perceptual features—have been extensively utilized. They have demonstrated promising performance in tasks such as expression classification and VA estimation\cite{zhang2018attention,stuhlsatz2011deep,lieskovska2021review}. These features can be extracted using pyAudioAnalysis\cite{giannakopoulos2015pyaudioanalysis}, and akin to visual features, deep learning has also been extensively applied in the extraction of acoustic characteristics.

In the previous iterations of the ABAW competition\cite{kollias2022abaw,jiang2022facial,jin2021multi,meng2022multi,zhang2022transformer,zhang2022continuous}, numerous teams have leveraged multimodal features. Meng et al.\cite{meng2022multi} proposed a model that utilized both auditory and visual features, ultimately securing the top position in the vocal affect estimation track. To fully exploit affective information in the wild, Zhang et al.\cite{zhang2022transformer} harnessed multimodal information from images, audio, and text, and proposed a unified multimodal framework for Action Unit detection and expression recognition. This framework achieved the highest scores in both tasks. These methodologies have convincingly demonstrated the efficacy of multimodal features in the realm of affective behavior analysis tasks.

\subsection{Multimodal Structure}
In early studies, \cite{zadeh2016multimodal,perez2013utterance} employed connected multimodal features to train Support Vector Machine (SVM) models, which were found to be deficient in effectively modeling multimodal information. Recent research on multimodal affective analysis has primarily utilized deep learning models to simulate the interaction of information within and between modalities. \cite{truong2019vistanet} developed a neural network model known as Visual Aspect Attention Network (VistaNet), which leverages visual information as a sentence-level alignment source. This multimodal architecture enables the model to focus more on these sentences when classifying emotions. Presently, the use of transformers for multimodal learning has become the mainstream in multimodal algorithms. In the domain of image-text matching, ALBEF\cite{li2021align}, to some extent inspired by the CLIP\cite{radford2021learning} model, has introduced the concept of multimodal contrastive learning into multimodal models, achieving a unification of multimodal contrastive learning and multimodal fusion learning.

In previous iterations of the ABAW\cite{kim2022facial,tallec2022multi,wang2022multi,zhang2022transformer}, have harnessed the transformer architecture and achieved remarkable results.

\section{Feature Extraction}
We fuse features from different neural networks to obtain more reliable emotional features and utilize these fused features for downstream tasks. By combining information from various feature extraction models such as ResNet and POSTER, we achieve a more comprehensive and accurate representation of emotions.
\subsection{Resnet-18}
ResNet\cite{he2016deep}(He et al. 2016) is a deep convolutional neural network (CNN) architecture designed to address the common issues of vanishing gradients and exploding gradients during the training of deep neural networks. Its core idea is the introduction of residual blocks, which incorporate skip connections, making the network easier to train. Instead of directly learning the mapping of each layer, the output of the residual block learns the residual between the input and output. This structure effectively mitigates the vanishing gradient problem. The pre-trained model of ResNet-18 can be used as a feature extractor, which first pretained on the MS-Celeb-1M\cite{guo2016ms}, and finally obtain a 512-dimensional visual feature vector. transforming images into high-dimensional feature vectors for use in other machine learning tasks, such as image retrieval and image similarity computation.
\subsection{POSTER}
The two-stream Pyramid crOss-fuSion TransformER network (POSTER)\cite{zheng2022poster} is a novel deep learning model specifically designed for video understanding tasks, such as action recognition and video classification. POSTER combines a pyramid structure with a two-stream architecture, leveraging cross-layer fusion and transformer networks to enhance video understanding performance. Extensive experimental results demonstrate that POSTER outperforms SOTA methods on RAF-DB with 92.05\%, AffectNet\cite{mollahosseini2017affectnet} (7 cls) with 67.31\%, and AffectNet (8cls) with 63.34\%, respectively . The dimension of the visual feature vectors is 768.
\subsection{POSTER2}
The proposed POSTER2\cite{mao2023poster} aims to improve upon the complex architecture of POSTER, which focus primarily on improvements in cross-fusion, dual-stream design, and multi-scale feature extraction.In cross-fusion, POSTER2 replaces traditional cross-attention mechanisms with window-based cross-attention mechanisms. The dual-stream design eliminates the branch from images to landmarks. Regarding multi-scale feature extraction, POSTER2 integrates multi-scale features of images and landmarks, replacing POSTER's pyramid design to alleviate computational burden. Experimental results demonstrate that POSTER2 achieves state-of-the-art Facial Expression Recognition (FER) performance on multiple benchmark datasets with minimal computational cost. It retains the same visual feature dimensionality of 768 as POSTER.
\subsection{FAU}
Facial Action Units (FAU), originally introduced by Ekman and Friesen\cite{ekman1978facial}, are strongly associated with the expression of emotions .In the fields of computer vision and human-computer interaction, Facial Action Units (FAUs) are widely employed in the development of facial expression analysis and facial recognition systems. By detecting and recognizing combinations of individual FAUs, it is possible to infer overall facial expressions and their emotional meanings. We utilize the OpenFace\cite{baltrusaitis2018openface} framework (Baltrusaitis et al., 2018) for FAU feature extraction, resulting in a 17-dimensional feature vector.

\section{Method}

\begin{figure}[tb]
  \centering
  \includegraphics[width=12cm]{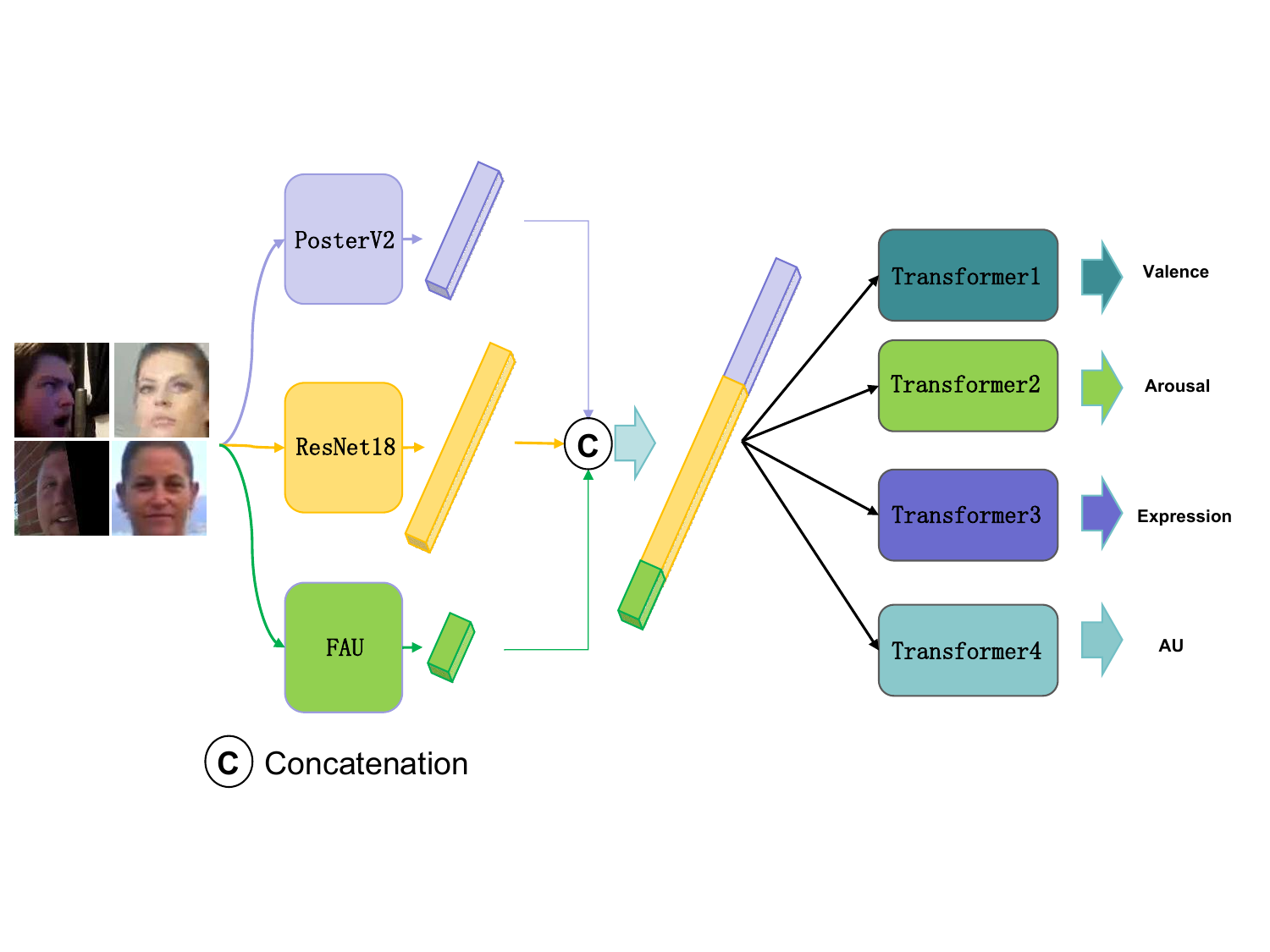}
  \caption{The overall framework of our proposed method. Visual Extractors contain EAC, ResNet18, POSTER, etc.The design of the transformer encoder is consistent with \cite{AshishVaswani2017AttentionIA}.
  }
  \label{fig:model}
\end{figure}

The 7th ABAW encompasses a total of two challenges, and we have participated in the first challenge. Drawing upon the design of classical transformer models, our entire process consists of four stages. As illustrated in \cref{fig:model} Initially, we utilize existing pre-trained models or toolkits to extract visual and auditory features from each frame of the video. Subsequently, each sequence of visual or auditory features is fed into an affine module to achieve features of the same dimension. Thirdly, these features are concatenated and then input into the transformer encoder to simulate temporal relationships. Lastly, the output of the encoder is fed into the output layer to obtain the corresponding outputs. The figure illustrates the overall framework of our proposed method.

\subsection{Affine Module}
In our experiments, the input consists of one or more visual features, but their feature dimensions often differ, and the discrepancies can be quite substantial. It can be observed that the EAC features span 2048 dimensions, while FAU has only 17 dimensions. We posit that excessively large dimensional disparities may diminish the effectiveness of the useful features. To address this, we have designed an affine module. For the first three challenges, we employ a linear layer to affinely transform features of varying dimensions to a uniform dimension. Furthermore, adhering to the setup of classical transformer\cite{AshishVaswani2017AttentionIA} models, we add positional encoding (PE) to each feature sequence to convey its contextual temporal information. The formula is as follows:

\begin{equation}
  \hat{f}_i = (Kf_i + c) + PE
  \label{eq:affine}
\end{equation}
where ${f_1,f_2,...,f_n}$ denote all the features, $n$ is the number of features. 

\subsection{Transformer Encoder}
Vectors from different feature extraction models may contain redundant or irrelevant information. To combine these features and construct a more suitable vector that retains more useful information for downstream classification tasks, we use a basic transformer encoder\cite{AshishVaswani2017AttentionIA}. In the context of classification tasks, the transformer\cite{AshishVaswani2017AttentionIA} model typically employs only the encoder part. The final vector is processed to capture the contexts and interdependencies of the data components. The output from the encoder is then typically passed through one or more dense layers to perform classification tasks such as Arousal-Valence and Action Units.

\begin{equation}
    \hat{f}=\left [ f_{1} ; f_{2};\cdots ; f_{n} \right ] 
\end{equation}
where $[ ; ; ]$ denotes the concatenation operation.

\begin{equation}
    T=TE(\hat{f})
\end{equation}
where T denotes the temporal feature.
\begin{equation}
    P=Mdl(T)
\end{equation}
where $Mdl$ represents the invoked model, and $p$ represents the resulting probability distribution.

\subsection{Loss Function}
For the distinct tasks within Challenge 1, we employ various loss functions tailored to the specific requirements of each task. We utilize the Mean Squared Error (MSE) and CCC loss to handle VA analysis, the Cross-Entropy loss for Expression Recognition, and a Weighted Asymmetric Loss for the Action Unit (AU) problem.
\begin{equation}
    L(p,\hat{p}) = \frac{1}{N}\sum_{i=1}^N (p_i-\hat{p}_i)^2
    \label{eq:mse_va}
\end{equation}
which $p_i$ and $\hat{p}_i$ is the label and prediction of valence or arousal, $N$ is the number of frames in a batch.
\begin{equation}
    L(p,\hat{p}) = -\sum_{i=1}^N \sum_{j=1}^C p_{ij} \log{\hat{p}_{ij}}
    \label{eq:crossentropy}
\end{equation}
which $p_{ij}$ and $\hat{p}_{ij}$ is the label and prediction of expression, $N$ is the number of frames in a batch and $C=8$ which denotes the number of expressions.
\begin{equation}
    L(p,\hat{p}) = -\frac{1}{N} \sum_{i=1}^N w_i[p_i\log{\hat{p}_i}+(1-p_i)\hat{p}_i\log{(1-\hat{p}_i)}]
    \label{eq:auloss}
\end{equation}
which $\hat{p}_i$, $p_i$ and $w_i$ are the prediction (occurrence probability), ground truth and weight of the $i^{th}$ AU. By the way, $w_i$ is defined by the occurrence rate of the $i^{th}$ AU in the whole training set.
\section{Experiments}
\subsection{Dataset}
The upcoming challenge will utilize the s-Aff-Wild2 database. s-Aff-Wild2 is the static version of the Aff-Wild2 database; it comprises frames selected from Aff-Wild2. In total, approximately 221K images will be employed, which include annotations on valence arousal; six basic expressions, along with a neutral state, plus an "other" category (encompassing emotional states not included in other categories); and 12 action units, namely AU1, AU2, AU4, AU6, AU7, AU10, AU12, AU15, AU23, AU24, AU25, and AU26.

Regarding the pre-trained visual feature extractors, two datasets are mentioned:

RAF-DB: A large-scale database containing around 30,000 facial images from numerous individuals, each image annotated approximately 40 times, followed by refinement using the EM algorithm to ensure the reliability of the annotations. 

AffectNet: A substantial facial expression recognition dataset comprising over one million images sourced from the internet. Approximately half of these images are manually annotated with seven discrete facial expressions, as well as the intensity levels of emotional value and arousal.

\subsection{Experimental Results}
The results, as presented in the table, indicate that the combination of POSTER2, ResNet18, and FAU features excels across various metrics, including emotional Valence-Arousal, expression recognition, and Action Unit (AU) detection, thereby underscoring their robustness. Specifically, the ResNet18 architecture has a notable influence on the recognition of facial expressions, while the FAU features are particularly impactful for the detection of Action Units.

\begin{table}
    \centering
    \setlength{\tabcolsep}{3mm}
    \begin{tabular}{l|c|c|c|c|c}
    \hline
    Features & Valence & Arousal & FER & AU & score \\
    \hline
    EAC & 0.414 & 0.425 & 0.249&0.433& 1.1015 \\
    POSTER & 0.439 & 0.347 & 0.247&0.423& 1.063 \\
    ResNet18 & 0.420 & 0.451 & 0.266&0.454& 1.1555 \\
    POSTER2 & 0.483 & 0.374 & 0.253 & 0.441 & 1.1775 \\
    ResNet18+Fau & 0.443 & 0.393 & 0.281&0.468& 1.167\\
    ResNet18+POSTER2 & 0.462 & 0.412 & 0.315&0.452& 1.204\\
    POSTER2+POSTER+EAC & 0.453 & 0.398 & 0.243&0.446& 1.1145\\
    ResNet18+POSTER2+FAU & 0.503 & 0.432 & 0.319&0.493& 1.2795\\
    \hline
    \end{tabular}
    \caption{The results on the validation set of Valence-Arousal Estimation, Facial Expression Recognition and Action Unit (AU) Detection with different features.}
    \label{tab:va}
\end{table}

\section{Conclusion}
In this paper, we introduce the methodologies presented at the 7th ABAW competition, encompassing three distinct sub-challenges: Valence Arousal (VA) estimation, Expression (Expr) classification, and Action Unit (AU) detection. We leveraged a robust suite of visual feature extractors and designed an affine module to standardize the varying impacts of individual feature sets. Our comprehensive experimental protocol demonstrates that our approach significantly surpasses the benchmarks, ensuring remarkable performance across all sub-challenges.
\bibliographystyle{splncs04}
\bibliography{main}
\end{document}